\documentclass[conference]{IEEEtran}
\IEEEoverridecommandlockouts
\usepackage[linesnumbered,ruled,vlined]{algorithm2e}
\usepackage{cite}
\usepackage{amsmath,amssymb,amsfonts}
\usepackage{graphicx}
\usepackage{url}
\usepackage{textcomp}
\usepackage{xcolor}
\def\BibTeX{{\rm B\kern-.05em{\sc i\kern-.025em b}\kern-.08em
    T\kern-.1667em\lower.7ex\hbox{E}\kern-.125emX}}
\begin{document}

\title{An Explainable AI Framework for Artificial Intelligence of Medical Things\\

}

\author{\IEEEauthorblockN{Al Amin$^{1}$, Kamrul Hasan$^{1}$, Saleh Zein-Sabatto$^{1}$, Deo Chimba$^{1}$, Imtiaz Ahmed$^{2}$, Tariqul Islam $^{3}$ }

\IEEEauthorblockA{$^{1}$ Tennessee State University, Nashville, TN, USA \\
$^{2}$Howard University, Washington, DC, USA \\
$^{3}$ Syracuse University, Syracuse, NY, USA \\
Email: $\lbrace$\textit{aamin2, mhasan1, mzein, dchimba}$\rbrace$@tnstate.edu,  $\lbrace$\textit{imtiaz.ahmed}$\rbrace$ @howard.edu, 
$\lbrace$\textit{mtislam}$\rbrace$ @syr.edu}}
\maketitle
\begin{abstract}
The healthcare industry has been revolutionized by the convergence of Artificial Intelligence of Medical Things (AIoMT), allowing advanced data-driven solutions to improve healthcare systems. With the increasing complexity of Artificial Intelligence (AI) models, the need for Explainable Artificial Intelligence (XAI) techniques become paramount, particularly in the medical domain, where transparent and interpretable decision-making becomes crucial. Therefore, in this work, we leverage a custom XAI framework, incorporating techniques such as Local Interpretable Model-Agnostic Explanations (LIME), SHapley Additive exPlanations (SHAP), and Gradient-weighted Class Activation Mapping (Grad-Cam), explicitly designed for the domain of AIoMT. The proposed framework enhances the effectiveness of strategic healthcare methods and aims to instill trust and promote understanding in AI-driven medical applications. Moreover, we utilize a majority voting technique that aggregates predictions from multiple convolutional neural networks (CNNs) and leverages their collective intelligence to make robust and accurate decisions in the healthcare system. Building upon this decision-making process, we apply the XAI framework to brain tumor detection as a use case demonstrating accurate and transparent diagnosis. Evaluation results underscore the exceptional performance of the XAI framework, achieving high precision, recall, and F1 scores with a training accuracy of 99\% and a validation accuracy of 98\%. Combining advanced XAI techniques with ensemble-based deep-learning (DL) methodologies allows for precise and reliable brain tumor diagnoses as an application of AIoMT. 
\end{abstract}
\section{Introduction}
Integrating electronic health (eHealth), intelligent healthcare systems and the Internet of Medical Things (IoMT) has led to AI-driven competent healthcare. However, traditional AI techniques face scalability and explainability challenges in modern healthcare networks. These sophisticated systems have amalgamated technology with healthcare provision, facilitating faster and more exact diagnosis and treatment of diverse health conditions, which also represents a significant advancement in improving patient outcomes and healthcare efficiency\cite{1,al2022adpt}. The healthcare industry is concerned about brain tumor diagnosis because millions of people are affected by brain tumors worldwide\cite{2}. It is not easy to diagnose accurately at different stages due to non-specific symptoms, imaging characteristics, and the presence of numerous tumor forms with various degrees of malignancy. Besides that, it is challenging to precisely define tumor boundaries and gauge their aggressiveness and genetic profiles. In addition, in different regions globally, the healthcare industries and AI methodologies encounter certain limitations in diagnostics due to the need for more adequately trained medical professionals and the unavailability of sophisticated medical equipment\cite{1}. AI applications require vast and well-annotated datasets to develop machine-learning models to diagnose brain tumors; however, gathering these datasets can be challenging due to data privacy issues, differences in imaging methods, and the rarity of specific tumor forms\cite{4}. In light of the difficulties encountered by the healthcare industry in achieving accurate and consistent diagnoses, this study proposes a series of technological innovations to overcome these obstacles:
\begin{itemize}
\item First, we design and propose an XAI framework specifically tailored for the field of Artificial Intelligence of Medical Things (AIoMT) to improve patient outcomes, autonomous diagnosis, and efficacy and accuracy in disease identification, such as brain tumors.
\item Second, we develop an XAI approach specific to AIoMT that amalgamates the maximum voting classifier (Ensemble) method and edge cloud-driven training, validation, and real-time evaluation. This integration aims to boost the reliability and accuracy of diagnoses, thereby aiding healthcare professionals in identifying brain tumors.
\item Finally, our research utilizes customized XAI techniques, encompassing methods like SHAP, LIME, and Grad-CAM, along with a custom ensemble algorithm to deliver accurate predictions with improved transparency, superior performance, and bias detection. As a result, medical professionals are equipped with a reliable toolset, fostering confidence in diagnosing brain tumors.
\end{itemize}

\section{Related Works}
In recent years, advances in AI  techniques have significantly improved the identification of brain tumors from magnetic resonance imaging (MRI) images, representing a significant advancement in the healthcare industry \cite{sharma2014brain,amin2019brain,nazir2021role,hemanth2019design}. This progress has been facilitated by integrating AI with electronic medical records (EMRs) and mobile health (mHealth) technologies \cite{saeed2015estimation}. In healthcare fields such as dermatology, radiology, and drug design, ML and XAI are becoming increasingly integral\cite{6,lee2021artificial,roy2022demystifying}. This research group deployed deep-learning approaches and transfer-learning architectures to identify and categorize brain tumors. Finally, they proposed the ensemble model's IVX16 and achieved a higher accuracy of around 96.94\%, but computational time is the major limitation\cite{8}. This research employs sigma filtering, adaptive thresholding, and region detection methods to identify brain tumors in MRI images, achieving 95\% precision. However, the potential for increased accuracy through dataset expansion and incorporating additional texture- and intensity-based features is a significant limitation\cite{10}. In another distinguished study, Arshia et al. harnessed transfer learning with three CNN architectures (i.e., GoogleNet, AlexNet, and VGGNet) and fine-tuned VGG16, achieving a commendable accuracy of $98.69\%$ in brain tumor detection and classification.
Nevertheless, the primary constraint of this research lies in the explainability of the DL model \cite{11}. In another pivotal study, researchers classified brain tumors using an explanation-driven, multi-input DL model. Using the SHAP and LIME methods, they provided an in-depth interpretation of the results, attaining a high level of accuracy of $94.64\%$\cite{13}. But authors mentioned that the better clinical issues, the research may be replicated and applied to other XAI algorithms such as GradCAM. Sarmad et al. used XAI in brain tumor detection research to highlight crucial image regions for concept prediction through gradients in the final convolutional layer. The main limitation is manipulating large amounts of data for real-life implementation\cite{14}. As highlighted in the mentioned challenges, this research adeptly navigates and resolves these complexities by leveraging a custom ensemble and XAI technique within a sophisticated edge-cloud-based framework.
\begin{figure}[t]
    \centering
    \includegraphics[scale=.6]{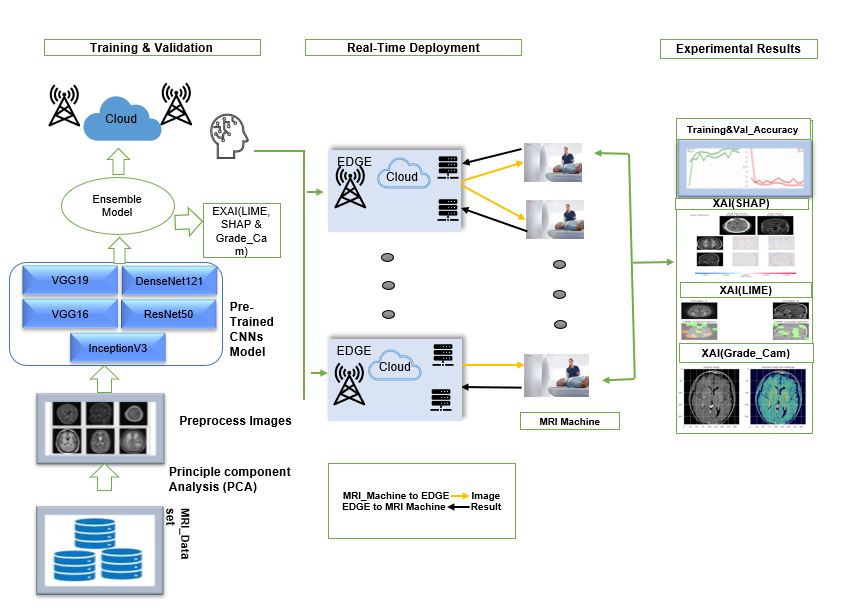}
     \caption{Illustration of the proposed artificial intelligence of medical things system.}
    \label{fig:my_label1}
    \vspace{-4mm}
\end{figure}

\section{System Architecture of AIoMT}
In this work, we consider an artificial intelligence of medical things system encompassed by several entities such as cloud, edge, medical devices, and monitoring systems. (as seen in Figure \ref{fig:my_label1}). This model expands and makes the transmission of computation between cloud and edge devices transparent by deploying a novel architecture that combines the benefits of cloud and edge platforms to facilitate the analysis of healthcare data from medical devices. We have enhanced our framework's security measures, ensuring data encryption and compliance with global privacy standards.

\subsection{AI-Driven Cloud-Edge based AIoMT System Description}
Our system model for cloud-edge integration is conceptually divided into three distinct phases, as depicted in Figure \ref{fig:my_label1}. 1) A First-stage cloud platform for training and validating MRI data in healthcare; 2) Edge computing will be incorporated with a cloud platform where healthcare applications will be deployed in real-time during the second stage; 3) Adding intelligence to both the cloud and the edge may significantly enhance the performance of healthcare applications, as will be explained to healthcare specialists in the final stages.

A designated cloud region securely stores the raw brain tumor MRI dataset in this proposed system. The cloud is integral in developing this system by providing a secure, scalable platform for storing and managing extensive MRI datasets and AI applications. It can handle large data volumes efficiently and provide easy access. We have also equipped it with powerful computational resources to leverage the intensive processing required for DL model training and validation. We employ Principal Component Analysis (PCA), a renowned statistical method for dimensionality reduction, to process this data. This step enables us to manage computational complexity and remove redundant information from the MRI scans. The ensemble model has been trained and validated using the MRI dataset, effectively aggregating the weights of the individual CNN models. This results in a robust and reliable model with enhanced prediction accuracy. XAI techniques, including SHAP, LIME, and Grad-CAM, have been employed to ensure that the decisions made by this model are transparent and interpretable.

The system architecture proposed has a two-tiered hierarchical structure. Multiple edge computing devices connected to various medical devices, such as MRI machines, comprise the first stratum. The second stratum is comprised of the central cloud platform that is linked to all edge servers. Each Edge device is directly related to one or more MRI machines. These Edge devices will acquire MRI scan images in real time. Because the EDGE server is connected to a cloud-based system where an ensemble model is trained on a large and diverse set of MRI images, the Edge server can now classify real-time MRI images into tumor and non-tumor categories. 

The final stage of this process is dedicated to experimental analysis, where we evaluate the overall system's effectiveness. The AI application's accuracy will be rigorously tested in this phase, and medical image analysis will be quantified. We will also generate visualizations using XAI techniques such as SHAP, LIME, and Grad-CAM. These visualizations serve a dual purpose: they enable us to comprehend and validate the model's decision-making process and provide interpretable insights into its findings for healthcare professionals, such as physicians and radiologists.

\subsection{Mathematical Formulation for the Proposed Maximum Voting Classifier Techniques}

This work considers $M$ different classification rules $h_M(X)$ for $X$ data, where $b \in M$. This technique is characteristic of ensemble methods, where numerous classifiers contribute their predictions, and the ultimate decision is based on the most common (or mode) vote\cite{17}
\begin{equation}
C_X = \text{mode} \{ h_1(X),\dots, h_M(X) \}.
\end{equation}
In other words, classify each $X$ value into the class that obtains the greatest number of classifications. This procedure can be applied to any number of classifiers, with certain classifiers having greater weight.\( w_j \) represents the weight assigned to each classifier \( j \). The weights can give certain classifiers more influence in the final decision. A majority vote classifier is typically comprised of votes from rules $h_1, h_2, \ldots, h_j$ as follows:
\begin{equation}
C_X = \arg\max_i \sum_{j=1}^B w_j I(h_j(X) = i).
\end{equation}

\begin{figure}[t]
    \centering
    \includegraphics[height=4.5cm, width=\linewidth]{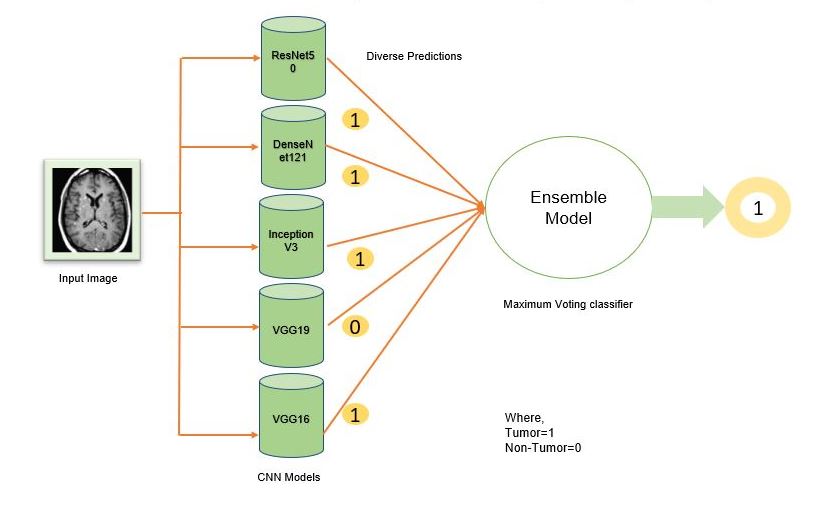}
    \caption{Maximum voting classifier diagram}
    \label{fig:my_label}
    \vspace{-4mm}
\end{figure}

This study utilizes binary cross-entropy to optimize the ensemble model loss function to quantify the difference between the actual class label and the model's predicted probabilities.
The binary cross-entropy loss for $n$ samples is calculated as \cite{17}:
\begin{equation}
L = -\frac{1}{n}\sum_{i=1}^{n}\left[y_i \log(\hat{y}_i) + (1 - y_i) \log(1 - \hat{y}_i)\right].
\end{equation}
This loss function is designed to push the model's predictions toward $1$ for the positive class and $0$ for the negative category. Here's a breakdown:
If $y_i$ is $1$ (positive class), the loss is $-\log(\hat{y}_i)$, which approaches $0$ as $\hat{y}_i$ approaches 1. Thus, the loss is small when the model correctly predicts a high probability for the positive class.
Equation 4 represents the loss when the class label is 1 (positive). It quantifies the difference between the model's prediction and the actual label, with the loss diminishing as the predicted probability for the positive class rises\cite{17}.
\begin{equation}
L_{y_i=1} = -\log(\hat{y}_i).
\end{equation}
If $y_i$ is 0 (negative class), the loss is $-\log(1 - \hat{y}_i)$, which approaches 0 as $\hat{y}_i$ approaches 0. Thus, the loss is slight when the model correctly predicts a low probability for the positive class (i.e., a high likelihood for the negative class). This equation represents the loss when the class label is 0 (negative). It also measures the difference between the model's prediction and the actual label, with the loss decreasing as the predicted likelihood of the hostile class rises.
\begin{equation}
L_{y_i=0} = -\log(1 - \hat{y}_i).
\end{equation}

% \subsection{Mathematical solutions for the Proposed Ensemble Model: }
The proposed custom ensemble model incorporated with the XAI framework can detect and explain the root cause of a particular disease to the doctor or medical staff for better healthcare outcomes. Therefore, base models and an input instance are combined with a maximum voting classifier to generate the output in the proposed ensemble model. The ensemble output is the most frequently predicted label, as determined by aggregating the 'votes' from the predictions of each base model, as shown in Figure\ref{fig:my_label}. In our research, we deploy a mechanism that selects the forecast with the highest score among the five function outputs. Given a set of base models $\{M_1, M_2, \dots, M_n\}$ and an input instance $x$, the output of the ensemble maximum voting classifier can be represented as:
\begin{equation}
y = \operatorname{mode}(f(M_1(x)), f(M_2(x)), \dots, f(M_n(x))),
\end{equation}

Where $f(Mi(x))$ is the predicted label of the input instance x by the i-th base model Mi, and mode() returns the most 
frequently occurring predicted label among all the base models' predictions. In other words, the ensemble model's 
output is the label that receives the highest number of votes from the base models.

%For our study purpose, we have used the ensemble-based mechanism, 
%\begin{equation}
%F(x) = \operatorname{argmax}  \{ f_1(x), %\dots, f_n(x) \}
%\end{equation}

\begin{algorithm}[h]
\scriptsize
\caption{Algorithmic Procedure of the Proposed XAI Framework for AIoMT:}\label{alg:tumorDetection}
\DontPrintSemicolon % Some LaTeX compilers require you to use \dontprintsemicolon instead
\KwData{all\_image\_sample}
\KwResult{final\_prediction}

\SetKwFunction{FMain}{Main}
\SetKwProg{Pf}{Function}{:}{}

\Pf{\FMain{all\_image\_sample}}{
    load\_image(all\_image\_sample)\;
    preprocessed\_images $\gets$ preprocess\_image\_for\_algorithms(all\_image\_sample)\;
    algorithms $\gets$ [VGG16, VGG19, Inception\_V3, ResNet50, DenseNet121]\;
    predictions $\gets$ [ ]\;

    \For{algorithm in algorithms}{
        prediction $\gets$ algorithm.predict(preprocessed\_images[algorithm])\;
        predictions.append(prediction)\;
        
    }end

    tumor\_votes $\gets$ 0\;
    non\_tumor\_votes $\gets$ 0\;

    \For{prediction in predictions}{
        \eIf{prediction == 'tumor'}{
            tumor\_votes $\gets$ tumor\_votes + 1\;
        }{
            non\_tumor\_votes $\gets$ non\_tumor\_votes + 1\;
        }end
    }end

    \eIf{tumor\_votes > non\_tumor\_votes}{
        final\_prediction $\gets$ 'tumor'\;
    }{
        final\_prediction $\gets$ 'non\_tumor'\;
    }
    \Return final\_prediction\;
    \textbf{print}("Final prediction:", final\_prediction)\;
}end

\end{algorithm}
\DecMargin{1em}

Algorithm \ref{alg:tumorDetection} implements an ensemble model for classifying MRI images of brain tumors. Initial loading and preprocessing of the raw image samples prepare them for algorithmic processing.

The following DL models are defined: VGG16, VGG19, Inception V3, ResNet50, and DenseNet121. We chose these image recognition algorithms because of their superior performance. Preprocessed image samples are fed to each algorithm in the collection to generate predictions. These predictions, indicating whether each image depicts a 'tumor' or a 'non-tumor," are then accumulated.

The algorithm \ref{alg:tumorDetection} then tallies the votes associated with 'tumor' and 'non-tumor' predictions. If the number of votes for tumor exceeds the number of votes for non-tumor, the ensemble model predicts that the final classification will be a tumor.

% \subsection{Advantages of this Algorithm}
The ensemble approach provides superior prediction accuracy by utilizing the best features of multiple DL models. Reducing bias and variance prevents overfitting, and the ensemble approach provides superior prediction accuracy. Integrating diverse models provides a broader perspective and enables a thorough analysis. Each model may capture different aspects or features in the data; hence, the ensemble will likely make a better-informed decision. As a result, the outcome of the developed algorithm \ref{alg:tumorDetection} significantly improved the medical personnel's confidence in disease detection and finding the root cause.

\subsection{The mathematical representation of XAI (LIME, SHAP, and Grad-CAM) in making determinations for the identification of brain tumors}
LIME: Medical imaging requires AI-driven interpretability. This work uses LIME to explain our classifier's brain tumor identification decision-making. LIME uses the original classifier to predict class labels for perturbed samples around a target instance. A localized weighted sample space is created by weighting samples by target similarity. LIME trains an interpretable, linear model in this space to approximate the original classifier's predictions near the target instance. The coefficients of this locally trained model provide categorization insights, frequently illustrated by masks highlighting essential image regions. Mathematically, the LIME-trained linear model is:
\begin{equation}
y' = \beta_0 + \beta_1 x_1' + \beta_2 x_2' + \ldots + \beta_n x_n' + \varepsilon
\end{equation}
Here, $y'$ is the predicted class label for the perturbed sample $x'$, $\beta_0$ is the intercept of the linear model, $\beta_1, \beta_2, \ldots, \beta_n$ are the coefficients of the linear model, $x_1', x_2', \ldots, x_n'$ are the features of the perturbed sample $x'$, and $\varepsilon$ is the error term. The trained model's coefficients $\beta_1, \beta_2, \ldots, \beta_n$ are then used to explain the original instance $x$ prediction.
The LIME optimization can be formulated as:
\begin{equation}
\min{f \in F} \sum w(x') (y' - f(x'))^2 + \Omega(f)
\end{equation}
Where $\mathcal{F}$ is the set of possible models, $f$ is the model being trained, $x'$ are the perturbed samples, $y'$ are the predicted class labels for the perturbed samples, $w(x')$ are the weights assigned to the perturbed samples, and $\Omega(f)$ is a complexity measure of the model, such as the number of non-zero coefficients in a linear model. The goal is to find the model $f$ that minimizes this objective function. The solution to this optimization problem explains the original instance.

SHAP: In our research, for detecting brain tumors, we utilize the XAI SHAP framework to elucidate the decision-making process of our DL-based ensemble model. Mathematically, the contribution of each feature, such as a pixel in an image, is quantified using Shapley values, defined as
$f(x)$, the SHAP value $\phi_i$ for feature $i$ is computed as:
\begin{equation}
\phi_i(f) = \sum_{S \subseteq N \setminus \{i\}} \frac{|S|!(|N|-|S|-1)!}{|N|!} [f(S \cup \{i\}) - f(S)]
\end{equation}
$N$ is the set of all features and $S$ is a subset of features excluding feature $i$. The term $f(S \cup \{i\}) - f(S)$ represents the contribution of feature $i$ when added to the subset $S$. By leveraging SHAP, we can understand how each feature (in this context, regions of an MRI image) influences the model's prediction, enabling a more transparent and interpretable AI-driven diagnosis.

Grad-CAM: Grad-CAM is a method that XAI uses to visually explain decisions made by ensemble models, particularly for applications like brain tumor detection. Grad-CAM highlights regions with the most significant impact on the model's prediction by superimposing a heatmap on the input image. In the context of brain tumor detection, this method provides invaluable insights by identifying specific regions in MRI images that the model considers indicative of the presence or absence of a tumor, thereby enhancing the transparency and reliability of the model's decision-making process. This mathematical equation represents the fundamental principle of Grad-CAM:
\begin{equation}
L^c = \sum_{i,j} \alpha^c_k A^k_{i,j}
\end{equation}
This equation represents the computation of the output for a specific class \( c \) by taking a weighted sum of values across different feature maps. Here, \( L^c \) denotes the output for class \( c \), while \( \alpha^c_k \) signifies the importance or weight of the \( k^{th} \) feature map for that class. The values within the feature map are represented by \( A^k_{i,j} \), where \( i \) and \( j \) are spatial indices. In essence, this equation captures how different features contribute to the classification decision for a given class by aggregating information from various feature maps, weighted by their respective importance.

\section{ Experimental Analysis}
\subsection{Dataset Description \& experimental result}
The Br35H dataset from the Kaggle web library was utilized in this work to identify brain tumors \cite{18}.  The dataset has two primary groups: tumors and non-tumors, with a total of 2590 and 500, respectively. The dataset was divided into two parts for the study's objectives: training and validation. The train-test split included 2472 samples for training purposes and 618 sample images to validate.
Table 1 shows the full classifier training and validation results for five different DL models and an ensemble model based on precision, recall, F1 score, training accuracy, validation accuracy, training loss, and validation loss in this study.
\begin{table}[h]
\label{tab1}
\caption{Comprehensive Assessment of Model Performance}
\resizebox{\linewidth}{!}{
\begin{tabular}{|l|l|l|l|l|l|l|l|}
\hline
Algorithms                                               & \begin{tabular}[c]{@{}l@{}}Precision\\  (\%)\end{tabular} & \begin{tabular}[c]{@{}l@{}}Recall\\  (\%)\end{tabular} & \begin{tabular}[c]{@{}l@{}}F1- \\ Score \\ (\%)\end{tabular} & \begin{tabular}[c]{@{}l@{}}Training \\ Accuracy\\    (\%)\end{tabular} & \begin{tabular}[c]{@{}l@{}}Training\\  loss (\%)\end{tabular} & \begin{tabular}[c]{@{}l@{}}Validation\\ Accuracy\\    (\%)\end{tabular} & \begin{tabular}[c]{@{}l@{}}Validation\\ loss   \\ (\%)\end{tabular} \\ \hline
VGG16                                                    & 95                                                        & 97                                                     & 96                                                           & 99                                                                     & 0.005                                                         & 98                                                                      & 0.06                                                                \\ \hline
VGG19                                                    & 96                                                        & 94                                                     & 95                                                           & 99                                                                     & 0.01                                                          & 97                                                                      & 0.07                                                                \\ \hline
InceptionV3                                              & 98                                                        & 97                                                     & 98                                                           & 98                                                                     & 0.05                                                          & 98                                                                      & 0.05                                                                \\ \hline

ResNet50                                                 & 94                                                        & 80                                                     & 85                                                           & 93                                                                     & 0.2                                                           & 93                                                                      & 0.24                                                                \\ \hline
DenseNet121                                              & 97                                                        & 94                                                     & 96                                                           & 98                                                                     & 0.05                                                          & 97                                                                      & 0.39                                                                \\ \hline
\begin{tabular}[c]{@{}l@{}}Ensemble\\ Model\end{tabular} & 98                                                        & 97                                                     & 99                                                           & 99                                                                     & 0.03                                                          & 98                                                                      & 0.04                                                                \\ \hline
\end{tabular}}
\end{table}

\subsection{Ensemble Model Performance: Confusion Matrix Analysis }

\begin{figure}[t]
    \centering
    \includegraphics[height=4.5cm, width=\linewidth]{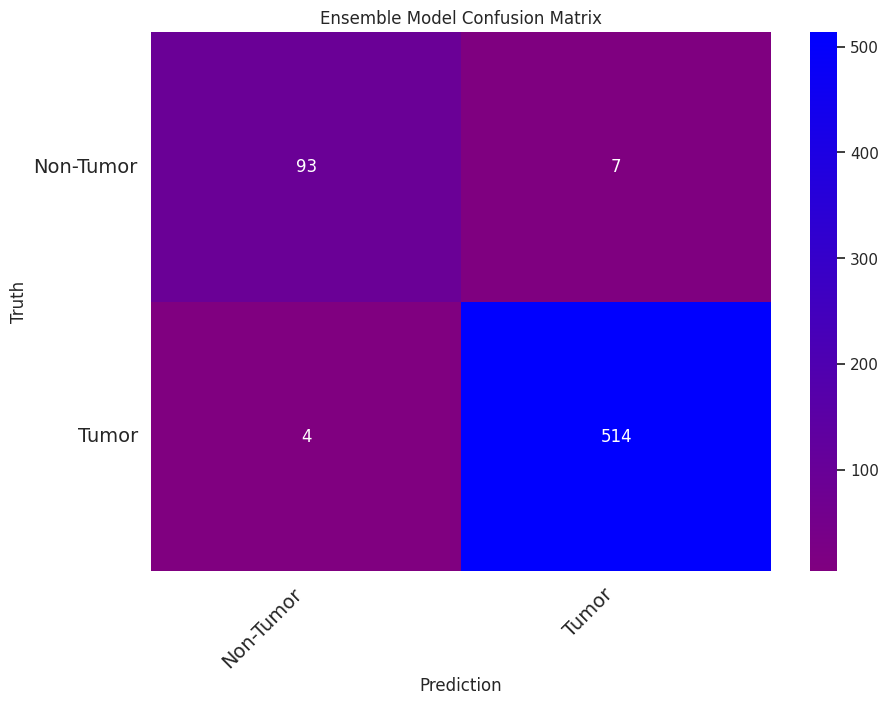}
    \caption{Confusion Matrix for Binary Classification(Ensemble model)}
    \label{fig:confusion1}
    \vspace{-4mm}
\end{figure}

Figure \ref{fig:confusion1} presents a confusion matrix used to assess the performance of an average ensemble model through the validation dataset. The matrix demonstrates the model’s capability to accurately predict tumors' existence or non-existence. The model correctly predicted non-tumor cases 93 times, as indicated by the true-negative rate. However, it inaccurately classified seven non-tumor cases as positive, representing the false-positive rate. Furthermore, the model misclassified four tumor cases as negative, marking the false-negative rate. Conversely, the model successfully identified 514 cases as positive tumors, illustrating the true-positive rate.

\subsection{Training and Validation Accuracy Curves for the Ensemble Model}
\begin{figure}[t]
    \centering
    \includegraphics[height=4cm, width=\linewidth]{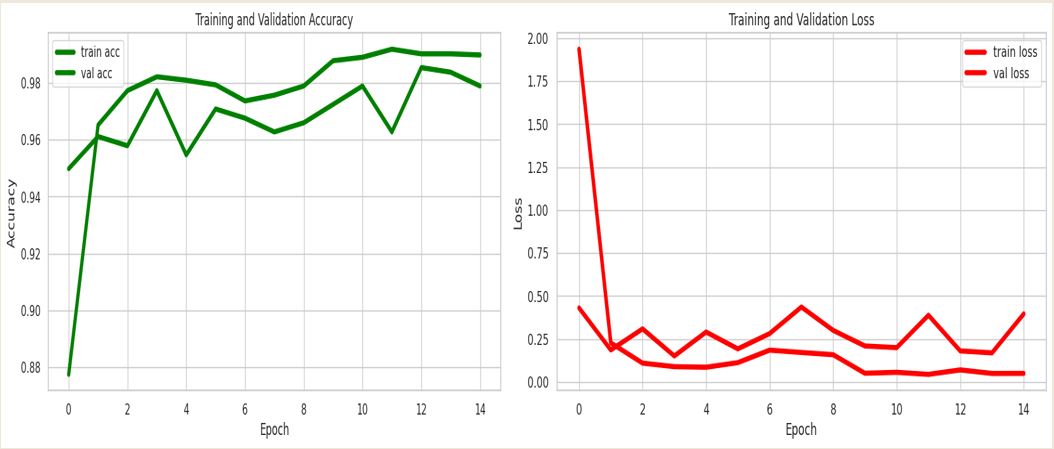}
    \caption{ Accuracy and loss graphs (training + validation) of a maximum ensemble approach. }
    \label{fig:validation}
    \vspace{-4mm}
\end{figure}

Figure \ref{fig:validation} for the proposed ensemble model, the training and validation accuracy curve indicates that the model is learning to classify brain tumors correctly over successive epochs during the training phase. On the other hand, the validation accuracy curve shows that the ensemble model generalizes to new, unseen data. The model validation accuracy is high and close to the training accuracy, which suggests that the model is not overfitting the training data and is effectively learning the underlying patterns.

\subsection{Receiver Operating Characteristic (ROC) Curve for the Ensemble Model in Brain Tumor Classification: }

\begin{figure}[t]
    \centering
    \includegraphics[height=4cm, width=\linewidth]{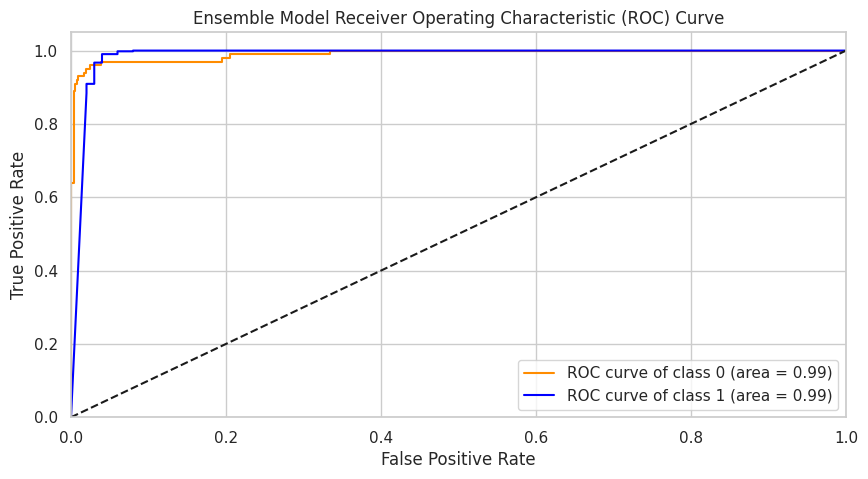}
    \caption{ Receiver Operating Characteristic (ROC) Curve for
ensemble approach}
    \label{fig:ROC}
    \vspace{-4mm}
\end{figure}

Figure \ref{fig:ROC} depicts the ensemble model's Receiver Operating Characteristic (ROC) curve. The ROC curve, a plot of the true-positive rate (sensitivity) against the false-positive rate (1-specificity), provides a comprehensive view of the model's performance across all possible classification thresholds. The area under the curve (AUC), serving as an aggregate measure of performance, is impressive, indicating the model's high capability to distinguish between positive (tumor) and negative (non-tumor) cases. The closer the curve follows the left-hand border and then the top edge of the ROC space, the more accurate the model is.

\subsection{Interpretable AI for Enhanced Brain Tumor Detection}

\begin{figure}[t]
    \centering
    \includegraphics[height=4.5cm, width=\linewidth]{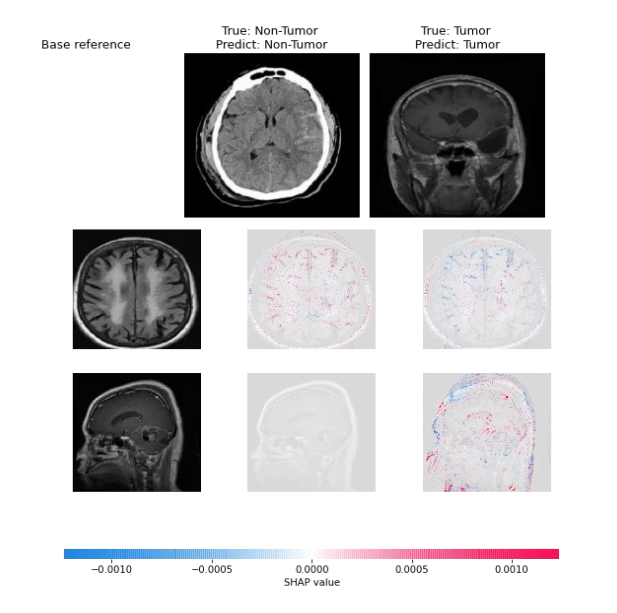}
    \caption{ SHAP explainable visualization. }
    \label{fig:SHAP}
    \vspace{-4mm}
\end{figure}

\begin{figure}[h]
    \centering
    \includegraphics[height=3cm, width=\linewidth]{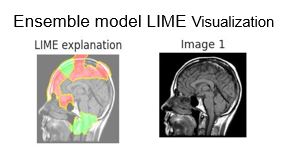}
    \caption{ LIME explainable visualization.}
    \label{fig:LIME}
    \vspace{-4mm}
\end{figure}

\begin{figure}[h]
    \centering
    \includegraphics[height=4.5cm, width=\linewidth]{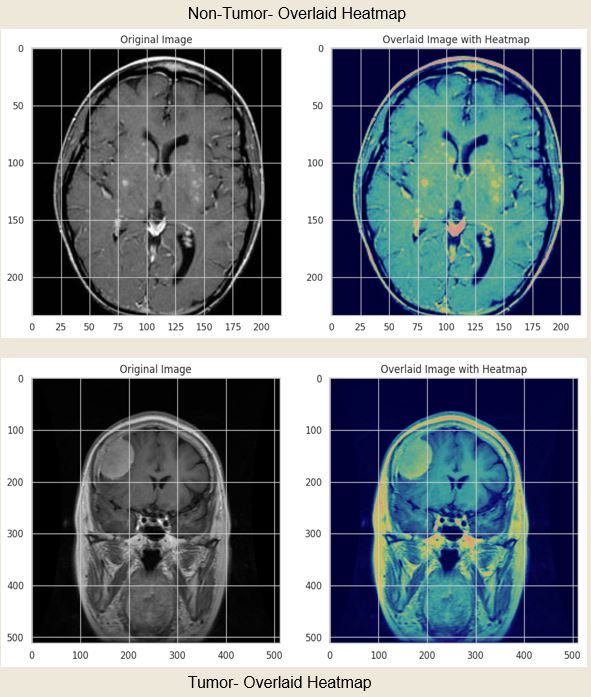}
    \caption{Grad-Cam explainable visualization.}
    \label{fig:Grade_Cam}
    \vspace{-4mm}
\end{figure}

As shown in Figures \ref{fig:SHAP}, \ref{fig:LIME}, and \ref{fig:Grade_Cam}, the proposed ensemble model, which employs proficient DL architectures, is coupled with XAI techniques, namely SHAP, LIME, and Grade-Cam, for clear visualization of model decisions. This method simplifies intricate AI decision-making for medical professionals and enhances the system's credibility. The color-coded visualizations improve interpretability by facilitating rapid comprehension of the factors influencing the model's predictions. These visualizations precisely depict influential features for a specific prediction, facilitating informed decisions. This technique bridges the gap between complex AI processes and practical healthcare applications, paving the way for the ethical use of AI in healthcare.

\section{Conclusion}
 This study introduces a robust framework employing XAI to implement AIoMT, particularly in detecting brain tumors. The framework utilizes an ensemble of a majority voting technique for enhanced decision-making in healthcare. Using custom XAI techniques such as LIME, SHAP, and Grad-CAM ensures that diagnoses are transparent and interpretable. The proposed novel ensemble model achieved significant results with high precision, recall, and F1 scores, as well as a training accuracy of $99\%$ and a validation accuracy of $98\%$. Our framework's integration with a cloud-edge system was instrumental in handling large datasets and computational complexities, making this AI-driven healthcare solution feasible for real-world applications. Despite the encouraging results, future work includes refining the framework to accommodate other types of tumors and medical conditions, exploring different ensemble techniques and XAI methods to augment the accuracy and interpretability, integrating real-time patient data, conducting longitudinal studies for evaluating long-term efficacy and adaptability, and addressing the critical ethical and privacy concerns inherent to AIoMT as the field continues to evolve.  

 \section*{Acknowledgment}
This material is based on work supported by the
National Science Foundation Award Numbers 2205773 and 2219658.

\bibliographystyle{ieeetr}
\bibliography{References.bib}
\end{document}